\documentclass[letterpaper, 10 pt, conference]{ieeeconf}  % Comment this line out if you need a4paper

\IEEEoverridecommandlockouts

\usepackage{cite}
\usepackage{amsmath,amssymb,amsfonts}
\usepackage{algorithmic}
\usepackage{graphicx}
\usepackage{textcomp}
\usepackage{xcolor}
\usepackage{booktabs}
\def\BibTeX{{\rm B\kern-.05em{\sc i\kern-.025em b}\kern-.08em
    T\kern-.1667em\lower.7ex\hbox{E}\kern-.125emX}}

\usepackage{tikz}
\usetikzlibrary{positioning,fit,backgrounds,calc,arrows.meta}
\usepackage{svg}
\usepackage{pgfplots}
\pgfplotsset{compat=1.18}

\pgfplotsset{
    paperlineplot/.style={
        width=0.96\linewidth,
        height=0.62\linewidth,
        axis line style={draw=black!45, line width=0.6pt},
        tick style={draw=black!45, line width=0.5pt},
        tick label style={font=\small, text=black!75},
        label style={font=\small, text=black!85},
        xminorticks=false,
        yminorticks=false,
        grid=major,
        xmajorgrids=false,
        ymajorgrids=true,
        grid style={draw=gray!18, line width=0.35pt},
        legend style={
            font=\small,
            draw=none,
            fill=none,
            text=black!80,
            at={(0.98,0.98)},
            anchor=north east,
            row sep=1pt
        },
        every axis plot/.style={
            line width=1.25pt
        },
        mark size=2.6pt,
        clip=false
    }
}

\title{\LARGE \bf
Phase-Aware Guidance Injection for Recurrent MAPPO in Assembly-Line Disruption Recovery
}

\author{Xin Huang$^{1,2}$, Yongcai Wang$^{1}$, Fengyi Zhang$^{3}$, Zhikun Tao$^{2,4}$, Yunjun Han$^{2,*}$, and Naiqi Wu$^{5}$%
\thanks{This work was supported by independent research project of the National Key Laboratory of Big Data and Decision (No. DSJ-ZZKY-2025-04);Public Computing Cloud, Renmin University of China, and the Blockchain Lab. School of Information, Renmin University of China.}%
\thanks{$^{1}$School of Information, Renmin University of China, Beijing 100872, China}
\thanks{$^{2}$State Key Laboratory of Multimodal Artificial Intelligence Systems, Institute of Automation, Chinese Academy of Sciences, Beijing 100190, China}
\thanks{$^{3}$The Information Science Academy, China Electronics Technology Group Corporation, Beijing 100043, China}
\thanks{$^{4}$School of Artificial Intelligence, University of Chinese Academy of Sciences, Beijing 100049, China}
\thanks{$^{5}$The Institute of Systems Engineering, Macau University of Science and Technology, 999078, Macau SAR, China}
\thanks{$^{*}$Corresponding author: Yunjun Han (yunjun.han@ia.ac.cn).}%
}

\begin{document}

\bstctlcite{IEEEexample:BSTcontrol}
\maketitle
\thispagestyle{empty}
\pagestyle{empty}

\begin{abstract}
Disruption recovery in industrial assembly lines requires timely decisions under machine faults, worker absence, and emergency orders. Existing methods either rely on rigid handcrafted recovery logic or learn adaptive policies that do not readily exploit heterogeneous external recovery knowledge at decision time to reduce abnormal recovery time (ART) and preserve on-time delivery (OTD). To address this gap, we propose a phase-aware guidance injection framework that augments a trained recurrent MAPPO (RMAPPO) scheduling policy through action-level bias during evaluation. The framework provides a unified decision-time interface for rule-based, replay-based, and online LLM-based guidance without retraining the actor, while activating intervention only during abnormal and recovery phases. Experiments on a custom AssemblyLineEnv show that high-quality rule guidance yields the strongest gains, replay-based guidance degrades smoothly under imperfect availability, and online LLM guidance still provides useful intermediate improvements. These results show that decision-time guidance injection can exploit heterogeneous recovery hints without redesigning the actor.
\end{abstract}

\section{INTRODUCTION}
Industrial assembly lines must continue operating under disruptions such as machine faults, worker absence, and emergency orders. Once such events occur, the system must rapidly shift from normal scheduling to recovery-oriented decision making so that delayed jobs can be cleared and delivery performance can be preserved. This setting is challenging because decisions are sequential and strongly coupled across stations: a local intervention can quickly propagate to downstream queues, workloads, and due-date pressure. As a result, disruption recovery requires not only adaptive scheduling policies but also timely use of recovery knowledge.

Traditional dispatching rules, reactive rescheduling schemes, and handcrafted heuristics are widely used in industrial scheduling because they are simple and interpretable \cite{vieira2003rescheduling,ouelhadj2009survey,sun2001dynamic}. However, they are often inflexible when disruption patterns vary over time. Reinforcement learning has increasingly been used to learn adaptive dispatching and scheduling behavior, and MARL further extends this idea to cooperative settings with multiple interacting decision makers \cite{zhang2020learning,yu2022mappo,liu2023deep,lv2025marlbreakdown}. Nevertheless, existing approaches mainly focus on rule design or policy learning itself, while giving limited attention to how external recovery knowledge can be injected into a trained policy during abnormal recovery.

Following the recovery mechanism considered in this work, we focus on three representative disruption categories: machine faults, worker absence, and emergency orders. These correspond to capacity loss, labor-resource reduction, and demand shock, respectively, and together define the abnormal-recovery setting studied in this paper.

To address this gap, we propose a guidance injection framework built on top of a trained recurrent MAPPO (RMAPPO) policy for disruption recovery in industrial assembly lines. Rather than replacing the learned actor or introducing a new training algorithm, the proposed method intervenes only at evaluation time by modifying action selection through action-level bias. The framework uses a phase-aware activation design: guidance is applied only when the system is under an active disruption or is still clearing delayed workload in the subsequent recovery phase. This design preserves the original policy structure while providing a unified interface for three guidance sources: rule-based guidance as a high-quality reference, replay-based stub guidance as a controllable proxy for imperfect guidance availability, and online LLM guidance as a practical source when handcrafted rules are unavailable or costly to maintain.

This paper makes three contributions: 1) it proposes a phase-aware decision-time guidance injection framework for abnormal recovery in industrial assembly lines without retraining the actor; 2) it places rule-based, replay-based, and online LLM-based guidance under the same action-level intervention interface, enabling controlled comparison under a fixed recurrent policy backbone; and 3) it reports a focused empirical study showing that the framework is effective not only under high-quality guidance, but also under imperfect replay-based guidance and practical online LLM guidance.
\section{RELATED WORK}
\subsection{Traditional methods for disruption recovery} 
Disruption recovery has long been an important topic in industrial scheduling and production systems. Early studies established general frameworks for manufacturing rescheduling and dynamic scheduling under disturbances such as machine breakdowns, urgent job insertion, and workforce variation \cite{vieira2003rescheduling,ouelhadj2009survey,sun2001dynamic}. Classical recovery methods often rely on dispatching rules, heuristic search, or problem-specific rescheduling logic, and more recent studies move closer to practical assembly settings by considering adaptive production rescheduling, worker allocation, and learning-based schedule repair under machine breakdowns \cite{adaptive2024rescheduling,gao2024digital,lv2025schedule}. These works improve recovery through handcrafted rules, optimization, or problem-specific repair logic. In contrast, our work studies how external recovery knowledge can be incorporated at decision time under a learned cooperative control policy.

\subsection{MARL for disruption recovery} Reinforcement learning has shown increasing promise in sequential decision problems under uncertainty, and it has been used to learn dispatching and control strategies directly from interaction data \cite{zhang2020learning}. Multi-agent reinforcement learning (MARL) extends this capability to cooperative systems with multiple interacting decision makers \cite{yu2022mappo}. In scheduling and control contexts, MARL has been applied to dynamic job-shop scheduling, flexible job-shop control, online layout planning, and machine-breakdown-aware shop-floor coordination \cite{liu2023deep,kaven2024multi,wan2025effective,inal2023dynamicjssp,lv2025marlbreakdown}. These methods improve adaptability by learning cooperative policies directly from environment interaction. In particular, PPO-style methods such as MAPPO have shown strong empirical performance in cooperative decision problems \cite{yu2022mappo}. However, existing MARL studies mainly focus on policy learning itself, while paying less attention to how external recovery knowledge can be incorporated into a trained actor at decision time. Our work instead examines how external recovery knowledge can be introduced at decision time for a trained recurrent cooperative policy, without redesigning the underlying policy backbone.

\subsection{External guidance for learned policies and disruption recovery} Another relevant line of work studies how external knowledge can shape learned policies. Classical ideas include reward and policy shaping \cite{ng1999policy,griffith2013policy}, constraint-based intervention such as shielding \cite{alshiekh2018safe}, invalid action masking \cite{huang2022invalid}, and rule-guided policy improvement \cite{tappler2025ruleguided}. Large language models have also been explored as auxiliary sources of high-level guidance for sequential decision systems and MARL pipelines \cite{chen2024efficient,li2025llm}. While these studies suggest that learned policies can benefit from structured external hints, our setting differs in two important respects. First, we focus on abnormal recovery in a cooperative assembly-line environment, where intervention must remain compatible with multi-agent coordination rather than merely filtering isolated unsafe actions. Second, we study a unified decision-time interface that allows heterogeneous recovery guidance to be compared under the same recurrent policy backbone.

Different from reward shaping, invalid action masking, shielding, and rule-guided training, our method does not modify rewards, actor parameters, or training procedures during evaluation. Rule, replay, and LLM sources are used only as external decision-time providers under a phase-aware soft-bias interface. These differences motivate the guidance injection framework introduced in the next section.
\section{PROBLEM FORMULATION}
We model disruption recovery in an industrial assembly line as a cooperative multi-agent sequential decision problem with $N$ stations. Each station is treated as an agent. At decision step $t$, agent $i$ receives a local observation $o_i^t\in\mathcal{O}_i$ and selects an action $a_i^t\in\mathcal{A}$. The joint action is denoted by $\mathbf{a}^t=(a_1^t,\ldots,a_N^t)$. The environment transition updates queue evolution, station workload, and overdue-job dynamics. Because local decisions propagate through downstream stations, the recovery process is naturally coupled and cooperative.

Following the recovery mechanism considered in this work, we consider three disruption types: machine faults, worker absence, and emergency orders, corresponding to capacity loss, labor-resource reduction, and demand shock. The recovery process is divided into an abnormal phase, during which a disruption is active, and a recovery phase, during which delayed workload is cleared.

The environment is a discrete scheduling abstraction of an assembly line, with detailed settings summarized in Table~\ref{tab:setup}. Agent $i$ observes a 12-dimensional local vector containing station availability, queue status, disruption identifiers, and time. The centralized critic uses the concatenation of all local observations. The guidance module uses only online state summaries, including phase, disruption type and station, queue lengths, total queue length, and estimated overdue count; it does not access future rewards, ART/OTD values, or hidden future events.

The recovery objective is to improve recovery efficiency while preserving delivery performance under disruptions. In our experiments, recovery efficiency is measured by abnormal recovery time (ART), while delivery performance is measured by on-time delivery (OTD).
\section{GUIDANCE INJECTION METHOD}

\begin{figure*}[t]
    \centering
    \includegraphics[width=\textwidth]{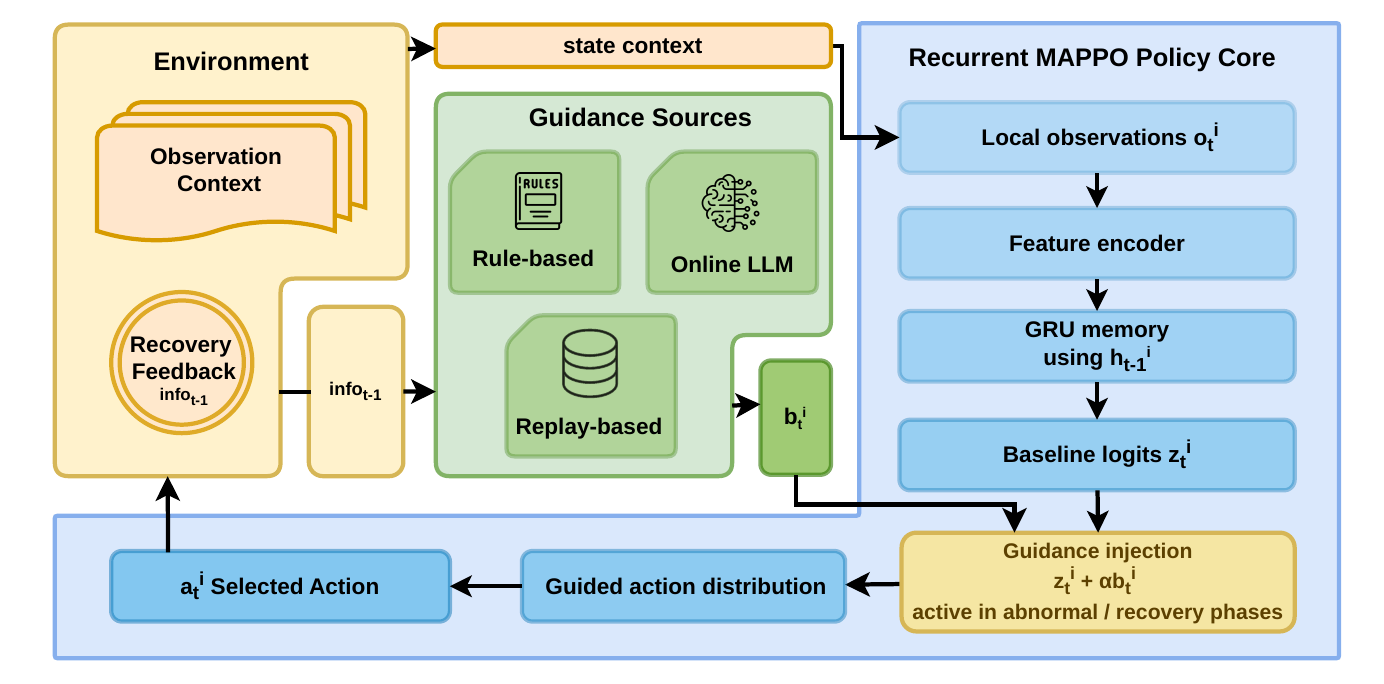}
    \caption{Overview of the proposed phase-aware guidance injection framework. The assembly-line environment provides observation context and recovery feedback to a recurrent policy backbone, which produces baseline action logits. Previous-step recovery feedback is converted by rule-based, replay-based, or online LLM-based guidance sources into the current-step bias signal, which is injected during abnormal and recovery phases to produce the guided action distribution.}
    \label{fig:framework}
\end{figure*}

We build the proposed method on top of a trained recurrent MAPPO (RMAPPO) actor and intervene only at evaluation time through external guidance. The framework modifies action selection so that the final decision reflects both the actor's learned preference and an external recovery hint. This design preserves the original actor structure while providing a unified interface for incorporating heterogeneous guidance sources.

The framework has three key properties. First, it is unified: rule-based, replay-based, and online LLM-based guidance are all injected through the same action-level interface. Second, it is controllable: the effect of guidance is explicitly adjusted by a guidance-strength parameter rather than being fixed. Third, it is practical: the same mechanism supports both high-quality offline guidance and noisy online guidance under a common evaluation pipeline.

\subsection{Baseline Recurrent MAPPO Actor and Logit-Level Guidance Injection}
Let $o_i^t$ denote the local observation of agent $i$ at step $t$, and let $h_i^{t-1}$ denote its recurrent hidden state. The trained recurrent MAPPO actor first encodes the local observation and updates the hidden state through a recurrent layer:
\begin{equation}
h_i^t=\mathrm{GRU}(g_{\phi}(o_i^t),h_i^{t-1}),
\end{equation}
where $g_{\phi}$ denotes the observation encoder and $h_i^t$ is the updated hidden representation for agent $i$. The action head then produces a logit vector over the discrete action space:
\begin{equation}
z_i^t=W h_i^t + c,
\end{equation}
where $W$ and $c$ are the action-head parameters, and $z_i^t\in\mathbb{R}^{|\mathcal{A}|}$. The corresponding baseline policy is
\begin{equation}
\pi_i^{\mathrm{base}}(a_i^t\mid o_i^t,h_i^{t-1})=\mathrm{softmax}(z_i^t)_{a_i^t}.
\end{equation}
We represent external guidance for agent $i$ at step $t$ as an action-bias vector $b_i^t\in\mathbb{R}^{|\mathcal{A}|}$, whose entries indicate which actions should be relatively encouraged or discouraged. Guidance is injected at the logit level by
\begin{equation}
\tilde{z}_i^t=z_i^t+\alpha b_i^t,
\end{equation}
where $\alpha\ge 0$ is a scalar guidance-strength parameter. The resulting guided policy is
\begin{equation}
\pi_i^{\mathrm{guide}}(a_i^t\mid o_i^t,h_i^{t-1},b_i^t)=\mathrm{softmax}(\tilde{z}_i^t)_{a_i^t}.
\end{equation}
When $\alpha=0$, the guided policy reduces exactly to the baseline policy. When $\alpha>0$, the external signal shifts the actor's action preference without modifying the actor architecture or retraining the policy. This logit-level intervention is the core mechanism that enables different guidance sources to be integrated through the same interface.

The implementation also supports an optional binary availability mask through the same action-selection layer. The mask is treated as a hard feasibility signal when it is supplied, whereas $b_i^t$ remains a soft preference signal. In the online LLM setting reported in this paper, the injected signal is restricted to soft action bias; no hard LLM-generated action mask is used.

\subsection{Phase-Aware Guidance for Abnormal Recovery}
External guidance is activated only during abnormal and recovery phases. The abnormal phase is detected when an active disruption is reported. The recovery phase starts immediately after the disruption terminates and continues until the estimated overdue count is zero for $K$ consecutive steps; the experiments use $K=5$. All remaining steps use the baseline learned policy without guidance. Guidance computed from the state summary after step $t-1$ is applied at step $t$, so the intervention does not use future trajectory information. This design concentrates recovery-oriented intervention on active disruption and backlog-clearing periods while leaving normal-operation decisions unchanged.

\subsection{Guidance Sources}
The proposed framework supports multiple guidance providers under the same injection interface. \textbf{Rule guidance} uses phase, disruption type and station, queue lengths, and overdue estimates to produce structured recovery bias. For example, it discourages parallel processing at a faulty station and encourages transfer or short-processing-time behavior where appropriate. It is used as a strong external reference, not as a new learned policy.

\textbf{Replay-based stub guidance} simulates imperfect guidance availability. A teacher database is constructed from rule-generated guidance using phase, disruption type, disruption station, queue lengths, and overdue estimate as keys. During evaluation, matched entries are retained with keep probability $p$ and otherwise dropped.

\textbf{Online LLM guidance} queries a large language model with the same online state summary. The prompt specifies the action set $\{\mathrm{EDD}, \mathrm{SPT}, \mathrm{parallel}, \mathrm{transfer}\}$ and requires JSON guidance. We parse only valid station-indexed bias vectors of length four with \texttt{temperature}=0; invalid outputs or request failures return empty guidance. The Qwen-plus online source is evaluated as soft action bias without hard LLM-generated masks.
\section{EXPERIMENTS}
\subsection{Experimental Setup}

We evaluate the proposed framework in a custom \texttt{AssemblyLineEnv}, which models a disrupted industrial assembly line as a cooperative multi-agent scheduling environment. The setting uses 5 stations, with each station treated as one agent and the joint policy executed in a fully cooperative manner. Each agent selects one action from the discrete set $\mathcal{A}=\{\mathrm{EDD},\mathrm{SPT},\mathrm{parallel},\mathrm{transfer}\}$, corresponding to earliest-due-date-first dispatching, shortest-processing-time-first dispatching, parallel processing behavior, and transfer/resequencing behavior, respectively. The environment includes three disruption types: machine fault, worker absence, and emergency order. During evaluation, the system may enter an abnormal phase when a disruption is active and a recovery phase after the disruption ends but delayed workload has not yet been fully cleared.

The shared RMAPPO reward is
\begin{equation}
r_t=\mathrm{throughput}_t-0.1Q_t-\widehat{O}_t-0.2C_t,
\end{equation}
where $Q_t$ is total queue length, $\widehat{O}_t$ is estimated overdue count, and $C_t$ is transfer count. The actor uses a shared policy, centralized critic, one GRU layer, hidden size 64, one MLP layer, learning rates $5\times10^{-4}$, PPO clip 0.2, 15 PPO epochs, one mini-batch, $\gamma=0.99$, GAE $\lambda=0.95$, entropy coefficient 0.01, value-loss coefficient 1, maximum gradient norm 10, and $10^5$ training steps.

We use two primary metrics: abnormal recovery time (ART) and on-time delivery (OTD). For each disruption event ending at step $e$, we search for the first step $t\ge e$ such that the estimated overdue count remains zero for $K$ consecutive steps, and define
\begin{equation}
\mathrm{ART}=t-e.
\end{equation}
If such a step does not exist before episode termination, we assign $\mathrm{ART}=T_{\max}-e$, where $T_{\max}$ is the maximum episode length. In all experiments, we use $K=5$. OTD is defined as
\begin{equation}
\mathrm{OTD}=\frac{N_{\mathrm{on\_time}}}{N_{\mathrm{on\_time}}+N_{\mathrm{late}}},
\end{equation}
where $N_{\mathrm{on\_time}}$ and $N_{\mathrm{late}}$ denote the numbers of on-time and late jobs, respectively.

All results reported in this paper use deterministic action selection. Unless otherwise stated, each episode is executed with a maximum of 200 steps. We evaluate across multiple environment seeds to assess stability under different disruption realizations. For the 5-station setting, we report 50-episode results across seeds 42, 43, and 44 for baseline RMAPPO, rule-guided RMAPPO, and online LLM-guided RMAPPO. For online guidance, we additionally use 20-episode runs for preliminary $\alpha$ scanning and refresh-frequency comparison under real online calls. We also verify that the guided evaluation pipeline reduces exactly to the baseline rollout when $\alpha=0$, which serves as a basic alignment check for fair no-guidance comparison.

To keep comparisons fair, all guided variants share the same trained recurrent MAPPO actor, environment configuration, episode horizon, and deterministic action-selection mode; the only difference is whether and how external guidance is injected during abnormal recovery. Performance changes can therefore be attributed to decision-time guidance rather than retraining or architecture modification.

\begin{table}[t]
\centering
\caption{Environment configuration and evaluation settings.}
\label{tab:setup}
\footnotesize
\setlength{\tabcolsep}{4pt}
\begin{tabular}{p{0.34\columnwidth} p{0.58\columnwidth}}
\hline
\textbf{Item} & \textbf{Setting} \\
\hline
Environment & \texttt{AssemblyLineEnv} \\
Stations & 5 \\
Agents & one agent per station \\
Action space & EDD, SPT, parallel, transfer \\
Disruption types & machine fault, worker absence, emergency order \\
Initial jobs / products & 40 jobs, 2 product types \\
Processing / due time & $U\{5,\ldots,14\}$ / $U\{50,\ldots,99\}$ \\
Disruption schedule & interval 30, duration 60 \\
Observation & 12-D local observation, 60-D shared observation \\
Max episode length & 200 \\
Evaluation mode & deterministic \\
Main metrics & ART, OTD \\
Default guidance strength & $\alpha = 4$ \\
Online LLM default & $\alpha = 4$, \texttt{llm\_every}=10 \\
\hline
\end{tabular}
\vspace{1mm}
\parbox{\columnwidth}{\footnotesize
\textit{Note:} ART = abnormal recovery time; OTD = on-time delivery;
\texttt{llm\_every} = the online LLM refresh interval (in decision steps).}
\end{table}

\subsection{Baselines}
We consider four settings under a unified evaluation pipeline. The three headline settings are reported in Table II, while replay-based guidance is analyzed in the robustness ablation. \emph{Baseline RMAPPO} uses the learned recurrent policy without external guidance. \emph{Rule-guided RMAPPO} augments the trained recurrent policy backbone with hand-designed recovery guidance and serves as the high-quality reference. \emph{Replay-based stub-guided RMAPPO} retrieves guidance from a teacher database generated by rule guidance, which allows us to simulate imperfect guidance availability in a controlled manner. \emph{Online LLM-guided RMAPPO} uses online LLM outputs in the form of soft action-bias signals. Unless otherwise noted, the reported comparisons focus on logit-level action-bias injection under the same recurrent policy backbone.
\subsection{Main Results}

\begin{table}[t]
\centering
\caption{Main results on the 5-station environment (50 episodes).}
\label{tab:main_results}
\footnotesize
\setlength{\tabcolsep}{5pt}
\begin{tabular}{c c c c c c c}
\hline
\textbf{Seed} & \textbf{B-ART} & \textbf{L-ART} & \textbf{R-ART} & \textbf{B-OTD} & \textbf{L-OTD} & \textbf{R-OTD} \\
\hline
42 & 6.99 & 5.85 & 4.18 & 0.8599 & 0.8829 & 0.9018 \\
43 & 9.64 & 6.38 & 5.82 & 0.8435 & 0.8694 & 0.8919 \\
44 & 7.68 & 6.33 & 5.82 & 0.8432 & 0.8593 & 0.8850 \\
\hline
Mean & 8.10 & 6.19 & 5.27 & 0.8489 & 0.8706 & 0.8929 \\
\hline
\end{tabular}
\vspace{1mm}
\parbox{\columnwidth}{\footnotesize
\textit{Note:} B = Baseline; L = online LLM guidance (implemented with Qwen-plus in this work);
R = rule-guided. ART = abnormal recovery time; OTD = on-time delivery.}
\end{table}

Table~\ref{tab:main_results} summarizes the 50-episode results on the 5-station environment across three seeds. Here, B, L, and R denote the baseline, online LLM-guided, and rule-guided settings, respectively. A clear pattern is observed: both online LLM guidance and rule guidance improve over the baseline, while rule guidance provides the strongest overall recovery performance. Averaged across seeds 42, 43, and 44, the baseline yields an ART of 8.10 and an OTD of 0.8489. Rule-guided RMAPPO reduces ART to 5.27 and improves OTD to 0.8929, corresponding to an average ART reduction of approximately 34.9\% and an OTD improvement of 4.41 percentage points. Online LLM guidance also improves over the baseline, reducing ART to 6.19 and improving OTD to 0.8706 on average, which corresponds to an ART reduction of approximately 23.7\% and an OTD improvement of 2.17 percentage points. These results show that high-quality external guidance can substantially strengthen the learned recovery policy.

The cross-seed pattern is consistent: rule guidance achieves the best ART and OTD on all seeds, and online LLM guidance remains better than the baseline on average. Since different seeds correspond to different disruption realizations and backlog propagation patterns, this consistency is important for evaluating recovery robustness. This suggests that guidance improves recovery behavior across different disruption realizations rather than fitting a single favorable trajectory.

\subsection{Ablation Study}
We conduct ablation analysis from two perspectives: guidance strength and guidance quality.

\begin{figure}[t]
    \centering
    \begin{tikzpicture}
\begin{axis}[
    paperlineplot,
    xmin=0.4, xmax=5.1,
    ymin=4.0, ymax=6.3,
    xtick={0.5,1,2,3,4,5},
    xlabel={Guidance strength $\alpha$},
    ylabel={Mean ART},
    legend style={
        font=\small,
        draw=none,
        fill=none,
        text=black!80,
        at={(0.98,0.98)},
        anchor=north east
    }
]

% main alpha curve -- 改成更柔和的青蓝色
\addplot[
    color=cyan!55!black,
    mark=*,
    mark options={fill=cyan!55!black,, draw=white, line width=0.6pt}
] coordinates {
    (0.5,6.04)
    (1.0,5.63)
    (2.0,4.87)
    (3.0,4.77)
    (4.0,4.18)
    (5.0,4.70)
};
\addlegendentry{Rule-guided ART}

% 右上角注释框
\node[
    anchor=north east,
    font=\scriptsize,
    text=black!75,
    draw=gray!25,
    rounded corners=1.5pt,
    fill=white,
    fill opacity=0.85,
    text opacity=1,
    inner sep=3pt
] at (axis description cs:0.98,0.82) {Best recovery at $\alpha=4$};

\end{axis}
\end{tikzpicture}
    \caption{Effect of guidance strength $\alpha$ under rule-guided RMAPPO on the 5-station environment (seed 42, 50 episodes). Recovery performance improves as $\alpha$ increases from 0.5 to 4, and slightly degrades when $\alpha$ is further increased to 5.}
    \label{fig:alpha_rule}
\end{figure}
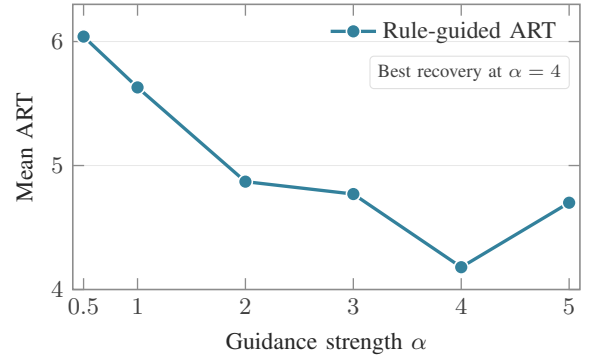

\textbf{Effect of guidance strength.} We first analyze the influence of the guidance-strength parameter $\alpha$ under rule guidance. As shown in Fig.~\ref{fig:alpha_rule}, performance improves steadily as $\alpha$ increases from 0.5 to 4. Specifically, ART decreases from 6.04 at $\alpha=0.5$ to 4.18 at $\alpha=4$, while OTD improves from 0.8732 to 0.9018. When $\alpha$ is further increased to 5, performance slightly degrades, with ART increasing to 4.70. This trend indicates that guidance must be strong enough to influence the learned actor, whereas overly strong guidance may reduce flexibility and slightly harm local adaptation. In our experiments, $\alpha=4$ provides the best overall trade-off and is therefore used as the default rule-guidance setting.

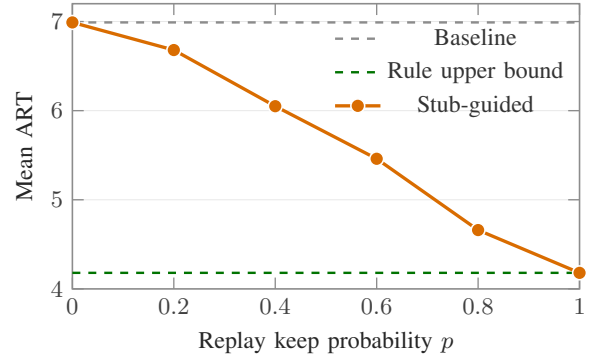
\begin{figure}[t]
    \centering
    \begin{tikzpicture}
\begin{axis}[
    paperlineplot,
    xmin=0, xmax=1,
    ymin=4.0, ymax=7.2,
    xtick={0,0.2,0.4,0.6,0.8,1.0},
    xlabel={Replay keep probability $p$},
    ylabel={Mean ART},
    legend style={
        font=\small,
        draw=none,
        fill=none,
        text=black!80,
        at={(0.98,0.98)},
        anchor=north east,
        xshift= 3pt,               
        yshift=-3pt           
    }
]

% baseline line
\addplot[
    color=black!45,
    dashed,
    line width=0.9pt
] coordinates {
    (0.0,6.99)
    (1.0,6.99)
};
\addlegendentry{Baseline}

% rule upper bound
\addplot[
    color=green!45!black,
    dashed,
    line width=0.9pt
] coordinates {
    (0.0,4.18)
    (1.0,4.18)
};
\addlegendentry{Rule upper bound}

% main stub curve
\addplot[
    color=orange!85!black,
    mark=*,
    mark options={fill=orange!85!black, draw=white, line width=0.6pt}
] coordinates {
    (0.0,6.99)
    (0.2,6.68)
    (0.4,6.05)
    (0.6,5.46)
    (0.8,4.66)
    (1.0,4.18)
};
\addlegendentry{Stub-guided}

\end{axis}
\end{tikzpicture}
    \caption{Robustness under replay-based stub guidance on the 5-station environment (seed 42, 50 episodes, $\alpha=4$). As replay availability increases, performance smoothly transitions from baseline-like behavior to rule-guided upper-bound performance.}
    \label{fig:stub_robustness}
\end{figure}

\textbf{Robustness under imperfect guidance.} We next study replay-based stub guidance generated from rule-based teacher signals. During evaluation, retrieved guidance is randomly retained with probability $p$, producing a controlled spectrum from no useful replay guidance ($p=0$) to full replay guidance ($p=1$). As shown in Fig.~\ref{fig:stub_robustness}, the degradation trend on seed 42 is smooth and monotonic: as $p$ increases from 0 to 1.0, ART decreases from 6.99 to 4.18, while OTD increases from 0.8599 to 0.9018. At $p=1.0$, the replay-guided result reaches the matched rule-guided result on the same seed. Together with the source-level comparison in Table~\ref{tab:main_results}, these results verify that each guidance source contributes useful recovery information: rule guidance provides the strongest improvement, replay-based guidance remains effective when available, and online LLM guidance offers stable intermediate gains.

These guidance sources represent different reliability--cost regimes: rule guidance is strong but manually specified, replay guidance reuses historical teacher traces, and online LLM guidance is flexible but uncertain and costly. Their compatibility with the same interface shows that the framework is not tied to one specific guidance form.

\subsection{Online LLM Guidance and Cost-Performance Trade-off}
We finally evaluate whether a practical online LLM can provide useful recovery guidance under the same injection interface. In all online experiments, online LLM guidance is injected as soft action bias without hard action masks; in our implementation, this source is instantiated with Qwen-plus. In 50-episode cross-seed evaluation with $\alpha=4$ and \texttt{llm\_every=10}, online LLM guidance consistently improves over the corresponding baseline on all three environment seeds. At the same time, the online results remain below the rule-guided reference, indicating that practical online guidance is useful but still weaker than high-quality handcrafted recovery knowledge.

The online prompt contains only the current phase, number of stations, disruption type and station, queue lengths, estimated overdue count, total queue length, and action dictionary. It excludes future rewards, ART, OTD, and ground-truth future recovery time. Malformed outputs are ignored by returning empty guidance.

We also compare different online refresh frequencies. To assess online cost--performance trade-offs rather than cross-seed headline performance, we additionally run a smaller 20-episode comparison on seed 42 with different refresh intervals. With $\alpha=4$ on seed 42, refreshing every 10 steps yields ART 6.93 and OTD 0.8773, while refreshing every 30 steps yields ART 7.10 and OTD 0.8798. Although the slower refresh slightly weakens recovery speed, it reduces the number of online LLM calls from 332 to 156. These results show a practical trade-off between recovery quality and online inference cost. The auxiliary statistics further suggest that the improvement from online LLM guidance is mainly associated with soft action-bias intervention rather than hard action masking. This matters in practice: a moderate but stable improvement may already be valuable if it is obtained through a unified interface without retraining the actor. The online LLM experiments therefore support the role of guidance injection as a general decision-time enhancement mechanism rather than a rule-only engineering trick.

\section{CONCLUSION}
This paper presented a guidance injection framework for disruption recovery in industrial assembly lines. By augmenting a trained recurrent MAPPO (RMAPPO) actor at evaluation time through action-level bias, the proposed method provides a unified and controllable way to incorporate external recovery knowledge without replacing the learned policy itself. Experiments on a custom AssemblyLineEnv show that high-quality external guidance substantially improves recovery performance. They also show that the guidance-strength parameter provides an explicit control knob over intervention intensity. Moreover, the framework remains effective under both imperfect replay-based guidance and practical online LLM guidance. These results suggest that decision-time guidance injection is a promising way to strengthen MARL-based recovery policies in disrupted industrial environments.

The current environment is a discrete scheduling abstraction and does not model high-fidelity robot motion, safety constraints, or real plant data. Future work will study improved online guidance quality, adaptive guidance scheduling, noisy phase detection, stronger baselines, and broader industrial recovery settings.

\bibliographystyle{IEEEtran}
\bibliography{refs}

@article{vieira2003rescheduling,
  author    = {Guilherme Ernani Vieira and Jeffrey W. Herrmann and Edward Lin},
  title     = {Rescheduling Manufacturing Systems: A Framework of Strategies, Policies, and Methods},
  journal   = {Journal of Scheduling},
  year      = {2003},
  volume    = {6},
  number    = {1},
  pages     = {39--62},
  doi       = {10.1023/A:1022235519958}
}

@article{ouelhadj2009survey,
  author    = {Djamila Ouelhadj and Sanja Petrovic},
  title     = {Survey of Dynamic Scheduling in Manufacturing Systems},
  journal   = {Journal of Scheduling},
  year      = {2009},
  volume    = {12},
  number    = {4},
  pages     = {417--431},
  doi       = {10.1007/s10951-008-0090-8}
}

@article{sun2001dynamic,
  author    = {Jie Sun and Dong Xue},
  title     = {A Dynamic Reactive Scheduling Mechanism for Responding to Changes of Production Orders and Manufacturing Resources},
  journal   = {Computers in Industry},
  year      = {2001},
  volume    = {46},
  number    = {2},
  pages     = {189--207},
  doi       = {10.1016/S0166-3615(01)00119-1}
}

@article{gao2024digital,
  author    = {Qinglin Gao and Jianhua Liu and Huiting Li and Cunbo Zhuang and Ziwen Liu},
  title     = {Digital Twin-Driven Dynamic Scheduling for the Assembly Workshop of Complex Products with Workers Allocation},
  journal   = {Robotics and Computer-Integrated Manufacturing},
  year      = {2024},
  volume    = {89},
  pages     = {102786},
  doi       = {10.1016/j.rcim.2024.102786}
}

@article{lv2025schedule,
  author    = {Lingling Lv and Jiaxin Fan and Chunjiang Zhang and Weiming Shen},
  title     = {Schedule Repair for Flexible Job Shops under Machine Breakdowns by Deep Reinforcement Learning},
  journal   = {Computers \& Industrial Engineering},
  year      = {2025},
  volume    = {207},
  pages     = {111256},
  doi       = {10.1016/j.cie.2025.111256}
}

@article{adaptive2024rescheduling,
  author    = {Andy J. Figueroa and Raul Poler and Beatriz Andres},
  title     = {Adaptive Production Rescheduling System for Managing Unforeseen Disruptions},
  journal   = {Mathematics},
  year      = {2024},
  volume    = {12},
  number    = {22},
  pages     = {3478},
  doi       = {10.3390/math12223478}
}

@inproceedings{yu2022mappo,
  author    = {Chao Yu and Akash Velu and Eugene Vinitsky and Jiaxuan Gao and Yu Wang and Alexandre Bayen and Yi Wu},
  title     = {The Surprising Effectiveness of {PPO} in Cooperative, Multi-Agent Games},
  booktitle = {Advances in Neural Information Processing Systems},
  year      = {2022},
  volume    = {35}
}

@inproceedings{zhang2020learning,
  author    = {Cong Zhang and Wen Song and Zhiguang Cao and Jie Zhang and Puay Siew Tan and Chi Xu},
  title     = {Learning to Dispatch for Job Shop Scheduling via Deep Reinforcement Learning},
  booktitle = {Advances in Neural Information Processing Systems},
  year      = {2020},
  volume    = {33}
}

@article{liu2023deep,
  author    = {Renke Liu and Rajesh Piplani and Carlos Toro},
  title     = {A Deep Multi-Agent Reinforcement Learning Approach to Solve Dynamic Job Shop Scheduling Problem},
  journal   = {Computers \& Operations Research},
  year      = {2023},
  volume    = {159},
  pages     = {106294},
  doi       = {10.1016/j.cor.2023.106294}
}

@article{kaven2024multi,
  author    = {Lea Kaven and Philipp Huke and Amon G{\"o}ppert and Robert H. Schmitt},
  title     = {Multi Agent Reinforcement Learning for Online Layout Planning and Scheduling in Flexible Assembly Systems},
  journal   = {Journal of Intelligent Manufacturing},
  year      = {2024},
  volume    = {35},
  number    = {8},
  pages     = {3917--3936},
  doi       = {10.1007/s10845-023-02309-8}
}

@article{wan2025effective,
  author    = {Lanjun Wan and Long Fu and Changyun Li and Keqin Li},
  title     = {An Effective Multi-Agent-Based Graph Reinforcement Learning Method for Solving Flexible Job Shop Scheduling Problem},
  journal   = {Engineering Applications of Artificial Intelligence},
  year      = {2025},
  volume    = {139},
  pages     = {109557},
  doi       = {10.1016/j.engappai.2024.109557}
}

@article{lv2025marlbreakdown,
  author    = {Lingling Lv and Jiaxin Fan and Chunjiang Zhang and Weiming Shen},
  title     = {A Multi-Agent Reinforcement Learning Based Scheduling Strategy for Flexible Job Shops under Machine Breakdowns},
  journal   = {Robotics and Computer-Integrated Manufacturing},
  year      = {2025},
  volume    = {93},
  pages     = {102923},
  doi       = {10.1016/j.rcim.2024.102923}
}

@article{inal2023dynamicjssp,
  author    = {Ali F{\i}rat {\.I}nal and {\c{C}}a{\u{g}}r{\i} Sel and Adnan Aktepe and Ahmet K{\"u}r{\c{s}}ad T{\"u}rker and S{\"u}leyman Ers{\"o}z},
  title     = {A Multi-Agent Reinforcement Learning Approach to the Dynamic Job-Shop Scheduling Problem},
  journal   = {Sustainability},
  year      = {2023},
  volume    = {15},
  number    = {10},
  pages     = {8262},
  doi       = {10.3390/su15108262}
}

@inproceedings{griffith2013policy,
  author    = {Shane Griffith and Kaushik Subramanian and Jonathan Scholz and Charles L. Isbell and Andrea L. Thomaz},
  title     = {Policy Shaping: Integrating Human Feedback with Reinforcement Learning},
  booktitle = {Advances in Neural Information Processing Systems},
  year      = {2013},
  volume    = {26}
}

@inproceedings{ng1999policy,
  author    = {Andrew Y. Ng and Daishi Harada and Stuart J. Russell},
  title     = {Policy Invariance Under Reward Transformations: Theory and Application to Reward Shaping},
  booktitle = {Proceedings of the Sixteenth International Conference on Machine Learning},
  year      = {1999},
  pages     = {278--287}
}

@inproceedings{alshiekh2018safe,
  author    = {Mohammed Alshiekh and Roderick Bloem and R{\"u}diger Ehlers and Bettina K{\"o}nighofer and Scott Niekum and Ufuk Topcu},
  title     = {Safe Reinforcement Learning via Shielding},
  booktitle = {Proceedings of the Thirty-Second AAAI Conference on Artificial Intelligence},
  year      = {2018},
  pages     = {2669--2678},
  doi       = {10.1609/aaai.v32i1.11797}
}

@inproceedings{huang2022invalid,
  author    = {Shengyi Huang and Santiago Onta{\~n}{\'o}n},
  title     = {A Closer Look at Invalid Action Masking in Policy Gradient Algorithms},
  booktitle = {Proceedings of the International FLAIRS Conference},
  year      = {2022},
  volume    = {35},
  pages     = {1025--1032}
}

@inproceedings{tappler2025ruleguided,
  author    = {Martin Tappler and Ignacio D. Lopez-Miguel and Sebastian Tschiatschek and Ezio Bartocci},
  title     = {Rule-Guided Reinforcement Learning Policy Evaluation and Improvement},
  booktitle = {Proceedings of the Thirty-Fourth International Joint Conference on Artificial Intelligence},
  year      = {2025},
  pages     = {6254--6262},
  doi       = {10.24963/ijcai.2025/696}
}

@article{li2025llm,
  author    = {Zhemin Li and Ruobing Zhang and Zhengming Wang and Zheng Xie and Yiping Song},
  title     = {{LLM}-Guided Decision-Making Toolkit for Multi-Agent Reinforcement Learning},
  journal   = {Neurocomputing},
  year      = {2025},
  volume    = {638},
  pages     = {130105},
  doi       = {10.1016/j.neucom.2025.130105}
}

@inproceedings{chen2024efficient,
  author    = {Dingyang Chen and Qi Zhang and Yinglun Zhu},
  title     = {Efficient Sequential Decision Making with Large Language Models},
  booktitle = {Proceedings of the 2024 Conference on Empirical Methods in Natural Language Processing},
  year      = {2024},
  pages     = {9071--9083}
}

\end{document}